\documentclass[letterpaper]{article}
\usepackage{aaai}
\usepackage{times}
\usepackage{helvet}
\usepackage{courier}
\usepackage{color,soul}
\usepackage{graphicx}
\graphicspath{{./figures/}} 
\usepackage{algorithm} 
\usepackage[noend]{algpseudocode} 
\begin{document}

\title{Cycle-of-Learning for Autonomous Systems from Human Interaction}

\author{Nicholas R. Waytowich\textsuperscript{1,2}, Vinicius G. Goecks\textsuperscript{3}, Vernon J. Lawhern\textsuperscript{1}\\
\textsuperscript{1}{US Army Research Laboratory}
\textsuperscript{2}{Columbia University}
\textsuperscript{3}{Texas A\&M University}\\
\texttt{nicholas.r.waytowich.civ@mail.mil},\\
\texttt{vinicius.goecks@tamu.edu}, \\
\texttt{vernon.j.lawhern.civ@mail.mil}
}

\maketitle
\begin{abstract}
\begin{quote}
We discuss different types of human-robot interaction paradigms in the context of training end-to-end reinforcement learning algorithms. We provide a taxonomy to categorize the types of human interaction and present our Cycle-of-Learning framework for autonomous systems that combines different human-interaction modalities with reinforcement learning. Two key concepts provided by our Cycle-of-Learning framework are how it handles the integration of the different human-interaction modalities (demonstration, intervention, and evaluation) and how to define the switching criteria between them.
\end{quote}
\end{abstract}

\section{Using Humans to Train Autonomous Systems}

Reinforcement learning (RL) has been successfully applied to solve challenging problems from playing video games to robotics. In simple scenarios, a reward function or model of the environment is typically available to the RL algorithm, and standard RL techniques can be applied. In real-world physical environments where reward functions are not available or are too intractable to design by hand, these standard RL methods often tend to fail. Using humans to train robotic systems is a natural way to overcome these burdens. 

There have been many examples in the field of human-robot interaction where human interaction is used to train autonomous systems in the context of end-to-end reinforcement learning. For example, imitation learning is an approach where demonstrations of a task provided from a human are used to initially train an autonomous system to learn a policy that ``imitates'' the actions of the human. A recent approach termed Human Centered Reinforcement learning (HCRL) trains autonomous systems using positive and negative feedback provided from a human trainer, showing promising strides for learning policies in the absence of a reward function. Approaches that learn control policies jointly with humans and autonomous systems acting together in a shared autonomy framework have also been developed.

Each of these human-interaction approaches, which have their own unique advantages and disadvantages, have mostly been utilized in isolation without taking into consideration the varying degrees of human involvement; the human is a valuable, but finite, source of information that can communicate information in many ways. A key question to consider is not only \textit{what} information the human should convey, but \textit{how} this information should be conveyed. A method for combining these different paradigms is needed to enable efficient learning from human interaction. 

\begin{figure*}[!t]
    \centering
    \includegraphics[width=.75\linewidth]{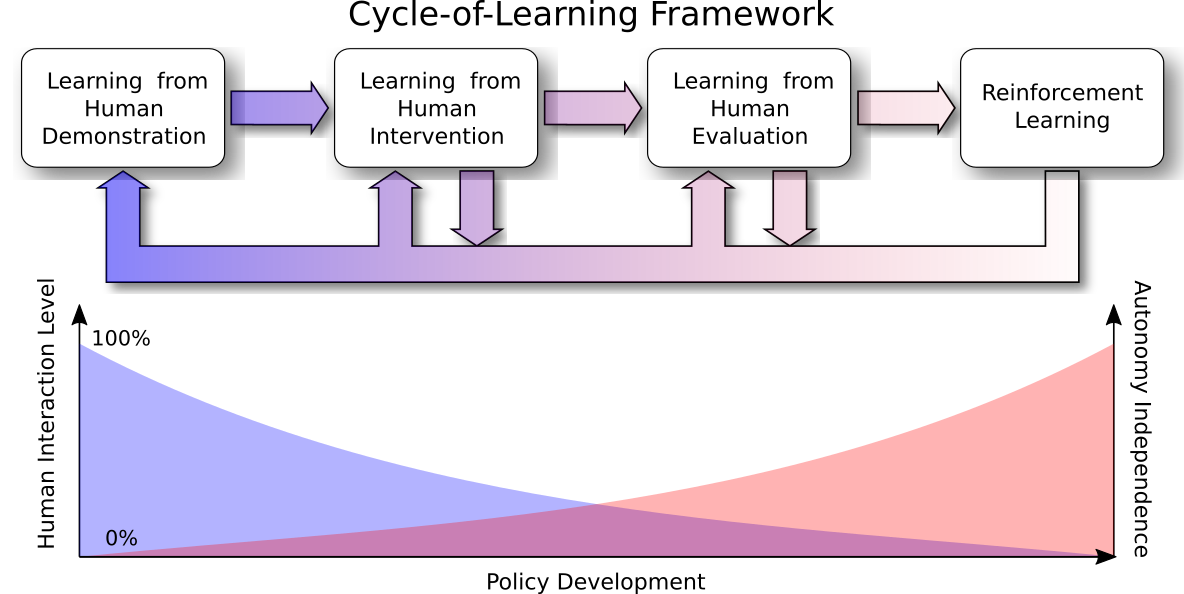}
    \caption{Cycle-of-Learning for Autonomous Systems from human Interaction: as the policy develops, the autonomy independence increases while the human interaction level decreases.}
    \label{fig:diagram}
\end{figure*}

In this paper, we present a new conceptual framework for using humans to train autonomous systems called the Cycle-of-Learning for Autonomous Systems from Human Interaction. This framework fuses together different human-interaction modalities into a single learning paradigm inspired by how humans teach other humans new tasks. We believe this intuitive concept should be employed whenever humans are interacting with autonomous systems. Our contributions in this paper are as follows: we first describe a taxonomy of learning from human interaction (with corresponding literature) and list the advantages and disadvantages of each learning modality. We then describe the Cycle-of-Learning (Figure~\ref{fig:diagram}), our conceptual framework for intuitively combining these interaction modalities for efficient learning for autonomous systems. We describe the different stages of our framework in the context of the degree of human involvement (and, conversely, the amount of autonomy of the system). We conclude with potential avenues for future research.   

\section{Types of Learning from Human Interaction}
There are many different ways a human may be used to train an autonomous system to perform a task, each of which can be broadly categorized by the modality of human-interaction. Here we present a taxonomy that categorizes most types of human interactions into one of three methodologies: Learning from human demonstrations (LFD), learning from human interventions (LFI) and learning from human evaluations (LFE). This taxonomy is partitioned based on the amount of control that the human or Autonomous system has during the learning process. In LFD the human is providing demonstrations and is in full control. In LFI, where the human occasionally intervenes, both the human and autonomous system share control. In LFE, where the human is providing evaluative feedback as the autonomous system performs a task, the autonomous systems is in control. We describe rec   ent research efforts for each of these categories and present their relative advantages and disadvantages. 

\subsection{Learning from Human Demonstrations}
Learning from human Demonstrations (LFD) is from a broad class of techniques called Imitation Learning (IL) where the aim is to train autonomous systems to mimic human behavior in a given task. In this interaction paradigm, the human takes the role of a demonstrator to provide examples of the task in the terms of sequences of states and actions. Using these demonstrations, the autonomous system learns a policy (a mapping from states to actions) that mimics the human demonstrations. 

There are many empirical successes of using imitation learning to train autonomous systems. For self-driving cars, Bojarski et al. successfully used IL to train a policy that mapped from front-facing camera images to steering wheel commands using around one hundred hours of human driving data \cite{Bojarski2016}. A similar approach was taken to train a small unmanned air system (sUAS) to navigate through forest trails, but in this case data was collected by a human following the trail \cite{Giusti2015}.

These results highlight both advantages and disadvantages of imitation learning approaches. IL can be used end-to-end to learn a new task, overcoming initial exploration of randomly-initialized learning algorithms while having better convergence properties because it trains on static datasets. However, the major drawback is that policy performance and generalization rely on the diversity and quality  and size of the training dataset --- often requiring large amounts of demonstrated behavior. 

Inverse Reinforcement Learning (IRL), also known as apprenticeship learning, is another form of learning from human demonstrations where a reward function is learned based on task demonstration. The idea is that during the demonstration, the human is using a policy that is optimal with respect to some internal reward function.
However, in IRL, the quality of the reward function learned is still dependent on the demonstration performance. In addition, IRL is fundamentally underdefined (degenerated), in a sense that different reward functions can lead to the same behavior \cite{Ng2000,Finn2016}. IRL focuses on learning the reward function, which is believed to be a more robust and transferable definition of the task when compared to the policy learned \cite{Ng2000}.

\subsection{Learning from Human Interventions}
Learning From human Interventions (LFI) is a much less explored method of human interaction for training autonomous systems. Simply put, the human acts as an overseer and intervenes (i.e. takes control) when the autonomous system is about to enter a catastrophic state. This is especially important for training embodied autonomous systems, such as quadrotors or ground robots, that are learning and interacting in a real environment where sub-optimal actions could lead to damage to the systems itself or the surrounding environment. Recently, this concept was formalized in a framework called learning from Human Interventions for safe Reinforcement Learning (HIRL) \cite{Saunders2017}. 
In HIRL, when the human observes the agent perform sub-optimal actions that could lead to a failure state, the human ``intervenes'' by blocking the current action and providing an alternative action. Additionally, a negative reward signal is given to the AI system during the intervention to learn from. 

By using humans to perform an initial state exploration and prevent catastrophic actions, LFI is a promising approach to solve the exploitation-exploration dilemma and increase safety in RL. Additionally, there is evidence that shared autonomy scenarios can also improve human performance \cite{Reddy2018}. However, LFI by itself does not scale to more complex environments where millions of observations are required for successful learning, increasing the required amount of human supervision \cite{Saunders2017}).

\subsection{Learning from Human Evaluations}
Another human-interaction modality that we consider is Learning From human Evaluations (LFE). In LFE, the human acts as a supervisor and provides real-time evaluations (or critiques) to interactively shape the behavior of the autonomous system.  There are several different approaches for how to provide the human feedback for LFE. One of the simplest approaches is fitting a reward function based on binarized feedback; for example, ``good'' vs ``bad'' actions, indicating positive and negative reward, respectively. Existing frameworks that use this approach include TAMER \cite{Knox2009,Warnell2018} and COACH \cite{MacGlashan2017}. Another approach asks the human to rank-order a given set of \textit{trajectories}, or short movie clips representing a sequence of state-action pairs, and learning a reward function in a relational manner based on human preferences \cite{Christiano2017}. Both approaches have been shown to be effective across a variety of domains and illustrates the utility of using human-centered reward shaping for shaping the policy of autonomous systems.

An advantage of LFE techniques is that they do not require the human to be able to perform demonstrations and only require an understanding of the task goal. However, if the time-scale of the autonomous system is faster than human reaction time, then it can be challenging for the autonomous system to attribute which actions correspond to the provided feedback.  In addition, the human reward signals are generally non-stationary and policy-dependent, i.e.: what was a good action in the past may not be a good action in the present depending on the humans perception of the autonomous system's policy.

\section{Proposed Cycle-of-Learning Framework}

The Cycle-of-Learning is a framework for training autonomous systems through human interaction that is based on the intuition on how a human would teach another human to perform a new task. For example, teachers conveying new concepts to their students proceed first by demonstrating the concept, intervening as needed while students are learning the concept, then providing critique after students have started to gain mastery of the concept. This process is repeated as new concepts are introduced. While extensive research has been conducted into each of these stages separately in the context of machine learning and robotics, to the best of our knowledge, a model incorporating each of these aspects into one learning framework has yet to be proposed. We believe such a framework will be important to fielding adaptable autonomous systems that can be trained on-the-fly to perform new behaviors depending on the task at hand, in a manner that does not require expert programming. 


\begin{algorithm}[!tb]
\caption{Cycle-of-Learning Framework}\label{alg:col}
\begin{algorithmic}[1]

\Procedure{Learning from Demonstration}{}
    \While {$SwitchingFunction_{LFD}:$}
        \State Collect human demonstration data $D_H$
        \State Train imitation learning policy $\pi_{\theta D_H}$
        \State Learn human reward function $R_H$ from $D_H$
    \EndWhile
\EndProcedure

\Procedure{Learning from Intervention}{}
    \While {$SwitchingFunction_{LFI}:$}
        \State Autonomous system performs the task
        \If {Human intervenes}:
            \State Collect human intervention data $D_I$
            \State Aggregate $D_H \gets D_H \cup D_I$
            \State Update imitation learning policy $\pi_{\theta D_H}$
            \State Update human reward function $R_H$
        \EndIf
        \State Compute intervention reward $R_I$
        \State Train critic $Q_{H}$ using $R_I$ and TD error
        \State Update policy $\pi_{\theta D_H}$ using policy gradient
    \EndWhile
\EndProcedure

\Procedure{Learning from Evaluation}{}
    \While {$SwitchingFunction_{LFE}:$}
        \State Collect human evaluation reward $r_H$
        \State Update critic $Q_{H}$ using $r_H$ and TD error
        \State Update policy $\pi_{\theta D_H}$ using policy gradient
        \State Update human reward function $R_H$ with $r_H$
    \EndWhile
\EndProcedure

\Procedure{Reinforcement Learning}{}
    \While {$SwitchingFunction_{RL}:$}
        \State Autonomous system performs the task
        \State Compute rewards using $R_H$
        \State Update critic $Q_{H}$ using $R_H$ and TD error
        \State Update policy $\pi_{\theta D_H}$ using policy gradient
    \EndWhile
\EndProcedure

\end{algorithmic}
\end{algorithm}

Under the proposed Cycle-of-Learning framework (Figure \ref{fig:diagram}), we start with LFD where a human would be asked to provide several demonstrations of the task. This demonstration data (observations received and actions taken) constitute the initial human dataset $D_H$. The dataset $D_H$ feeds an imitation learning algorithm to be trained via supervised learning, resulting in the policy $\pi_{\theta D_H}$. In parallel to the policy training, the dataset $D_H$ is used by an Inverse Reinforcement Learning (IRL) algorithm to infer the reward function $R_H$ used by the human while demonstrating the task (Algorithm \ref{alg:col}, line 1).

On LFI (Algorithm \ref{alg:col}, line 6) the autonomous system performs the task according to the policy $\pi_{\theta D_H}$. During the task the human is able to intervene by taking over the control of the autonomous system, perhaps to avoid catastrophic failure, and provides more demonstrations during the intervention. This new intervention dataset $D_I$ is aggregated to the previous human dataset $D_H$. Using this augmented dataset, the policy $\pi_{\theta D_H}$ and the reward model $R_H$ are updated. An intervention reward $R_I$ is computed based on the degree of the intervention. The reward signal $R_I$ and the temporal-difference (TD) error associated with it are used to train a value function $Q_H$ (the critic) and evaluate the actions taken by the actor. At this point, the policy $\pi_{\theta D_H}$ is updated using actor-critic policy gradient methods. 

After the human demonstration and intervention stages, the human assumes the role of a supervisor who evaluates the autonomous system actions through a reward signal $r_H$ --- Learning from Evaluation (LFE, Algorithm \ref{alg:col}, line 17). Similarly to the LFI stage, the reward signal $r_I$ and the TD error associated with it are used to update the critic $Q_H$ and the policy $\pi_{\theta D_H}$. The reward model $R_H$ is also updated according to the signal $r_H$ plus the observations and actions associated with it.

The final stage is pure Reinforcement Learning (RL). The autonomous system performs the task and its performance is evaluated using the learned reward model $R_H$ (Algorithm \ref{alg:col}, line 23). Similar to the LFI and LFE stages, the reward signal $R_H$ and the TD error associated with it are used to update the critic $Q_H$ and the policy $\pi_{\theta D_H}$. This sequential process is repeated as new tasks are introduced. 

\subsection{Integrating and switching between interaction modalities}
Two key concepts of the Cycle-of-Learning framework are how to handle the integration of the learned models from the different interaction modalities (demonstration, intervention, and evaluation) and how to define the criteria to switch between them. First, to integrate the different interaction modalities, we propose using an actor-critic architecture \cite{Sutton1999}: initially training only the actor, and later adding the critic. Training the actor first allows the framework to leverage the initial demonstration and intervention provided by the human. The critic is then trained as the human assumes the role of supervisor. After enough human demonstration data has been collected we can infer a reward function through IRL. At the end, the actor and critic are combined on a standard actor-critic reinforcement learning architecture driven by the learned reward model.

Second, we propose different concepts to define a criteria to switch between interaction modalities: performance metrics, data modality limitation, and advantage functions.
\textit{Performance metrics: }A pre-defined performance metric can be used to indicate when to switch modalities once the policy reaches a certain level. Alternatively, the human interacting with the system could manually switch between different interaction modalities as s/he observes that the autonomous system performance is not increasing.
\textit{Data modality limitations: }Depending on the task, there can be a limited amount of demonstration, intervention, or evaluation that can be provided by humans. In this case, the framework switches between modalities according to data availability.
\textit{Advantage functions: }After training the reward model $R_H$, advantages $A(s,a)$ (the difference between the state-action value function $Q(s,a)$ and the state value function $V(s)$, which compares the expected return of a given state-action pair to the expected return on that state) can be computed and used for expected return comparison between human and autonomous systems actions. With this information, the framework could switch interaction modalities whenever the advantage function of the autonomous system surpasses the advantage function of the human.  These, as well as other potential concepts for modality switching, need to be further investigated and can be adapted to meet task requirements.

\section{Discussion and Conclusions}

This paper presents the Cycle-of-Learning framework, envisioning the integration between different human-interaction modalities and reinforcement learning algorithms in an efficient manner. The main contributions of this work are (1) the formalization of the underlying learning architecture --- first leveraging human demonstrations and interventions to train an actor policy and reward model, then gradually moving to training a critic and fine-tuning the reward model based on the same interventions and additional evaluations, to finally combining these different parts on an actor-critic architecture driven by the learned reward model and optimized by a reinforcement learning algorithm --- and (2) the switching between these human-interaction modalities based on performance metrics, data modality limitations, and/or advantage functions.

We believe the proposed Cycle-of-Learning framework is most suitable for robotic applications, where both human and autonomous system resources are valuable and finite. As future work, it is planned to demonstrate these techniques on a human-sUAS (small unmanned air system) scenario.

\bibliographystyle{aaai}

\bibliography{RL.bib}

\begin{thebibliography}{}

\bibitem[\protect\citeauthoryear{Bojarski \bgroup et al\mbox.\egroup
  }{2016}]{Bojarski2016}
Bojarski, M.; {Del Testa}, D.; Dworakowski, D.; Firner, B.; Flepp, B.; Goyal,
  P.; Jackel, L.~D.; Monfort, M.; Muller, U.; Zhang, J.; Zhang, X.; Zhao, J.;
  and Zieba, K.
\newblock 2016.
\newblock {End to End Learning for Self-Driving Cars}.
\newblock  1--9.

\bibitem[\protect\citeauthoryear{Christiano \bgroup et al\mbox.\egroup
  }{2017}]{Christiano2017}
Christiano, P.; Leike, J.; Brown, T.~B.; Martic, M.; Legg, S.; and Amodei, D.
\newblock 2017.
\newblock {Deep reinforcement learning from human preferences}.
\newblock {\em arXiv}.

\bibitem[\protect\citeauthoryear{Finn, Levine, and Abbeel}{2016}]{Finn2016}
Finn, C.; Levine, S.; and Abbeel, P.
\newblock 2016.
\newblock {Guided Cost Learning: Deep Inverse Optimal Control via Policy
  Optimization}.
\newblock {\em Icml 2016} 48(2000).

\bibitem[\protect\citeauthoryear{Giusti \bgroup et al\mbox.\egroup
  }{2015}]{Giusti2015}
Giusti, A.; Guzzi, J.; Cire, D.~C.; He, F.-l.; Rodr{\'{i}}guez, J.~P.; Fontana,
  F.; Faessler, M.; Forster, C.; Schmidhuber, J.; Caro, G.~D.; Scaramuzza, D.;
  and Gambardella, L.~M.
\newblock 2015.
\newblock {A Machine Learning Approach to Visual Perception of Forest Trails
  for Mobile Robots}.
\newblock 3766(c):1--7.

\bibitem[\protect\citeauthoryear{Knox and Stone}{2009}]{Knox2009}
Knox, W.~B., and Stone, P.
\newblock 2009.
\newblock {Interactively Shaping Agents via Human Reinforcement: The TAMER
  Framework}.
\newblock In {\em International Conference on Knowledge Capture}.

\bibitem[\protect\citeauthoryear{MacGlashan \bgroup et al\mbox.\egroup
  }{2017}]{MacGlashan2017}
MacGlashan, J.; Ho, M.~K.; Loftin, R.~T.; Peng, B.; Roberts, D.~L.; Taylor,
  M.~E.; and Littman, M.~L.
\newblock 2017.
\newblock Interactive learning from policy-dependent human feedback.
\newblock {\em CoRR} abs/1701.06049.

\bibitem[\protect\citeauthoryear{Ng and Russell}{2000}]{Ng2000}
Ng, A., and Russell, S.
\newblock 2000.
\newblock {Algorithms for inverse reinforcement learning}.
\newblock {\em Proceedings of the Seventeenth International Conference on
  Machine Learning} 0:663--670.

\bibitem[\protect\citeauthoryear{Reddy, Levine, and Dragan}{2018}]{Reddy2018}
Reddy, S.; Levine, S.; and Dragan, A.~D.
\newblock 2018.
\newblock Shared autonomy via deep reinforcement learning.
\newblock {\em CoRR} abs/1802.01744.

\bibitem[\protect\citeauthoryear{Saunders \bgroup et al\mbox.\egroup
  }{2017}]{Saunders2017}
Saunders, W.; Sastry, G.; Stuhlm{\"{u}}ller, A.; and Evans, O.
\newblock 2017.
\newblock Trial without error: Towards safe reinforcement learning via human
  intervention.
\newblock {\em CoRR} abs/1707.05173.

\bibitem[\protect\citeauthoryear{Sutton \bgroup et al\mbox.\egroup
  }{1999}]{Sutton1999}
Sutton, R.~S.; Mcallester, D.; Singh, S.; and Mansour, Y.
\newblock 1999.
\newblock {Policy Gradient Methods for Reinforcement Learning with Function
  Approximation}.
\newblock {\em In Advances in Neural Information Processing Systems 12}
  1057--1063.

\bibitem[\protect\citeauthoryear{Warnell \bgroup et al\mbox.\egroup
  }{2018}]{Warnell2018}
Warnell, G.; Waytowich, N.; Lawhern, V.; and Stone, P.
\newblock 2018.
\newblock Deep tamer: Interactive agent shaping in high-dimensional state
  spaces.
\newblock {\em AAAI Conference on Artificial Intelligence}  1545--1553.

\end{thebibliography}

\end{document}